\documentclass[conference]{IEEEtran}
\IEEEoverridecommandlockouts
\usepackage{cite}
\usepackage{amsmath,amssymb,amsfonts}
\usepackage{algorithmic}
\usepackage{graphicx}
\usepackage{textcomp}
\usepackage{xcolor}
\usepackage{caption}
\usepackage{subcaption}
\usepackage{algorithm2e}
\usepackage{url}
\usepackage{lscape}
\usepackage{hyperref}

\usepackage{booktabs} 
\usepackage{graphicx}
\SetKwInOut{Parameter}{Parameter}

\newcommand*\WATCH{\texttt{WATCH}}

\def\BibTeX{{\rm B\kern-.05em{\sc i\kern-.025em b}\kern-.08em
    T\kern-.1667em\lower.7ex\hbox{E}\kern-.125emX}}
\begin{document}
\title{WATCH: Wasserstein Change Point Detection for High-Dimensional Time Series Data}
\author{\centering
\IEEEauthorblockN{Kamil Faber}
\IEEEauthorblockA{\textit{Institute of Computer Science} \\
\textit{AGH University of Science and Technology}\\
Krakow, Poland \\
kfaber@agh.edu.pl}
\and
\IEEEauthorblockN{Roberto Corizzo}
\IEEEauthorblockA{\textit{Department of Computer Science} \\
\textit{American University}\\
Washington, DC, USA \\
rcorizzo@american.edu}
\and
 \IEEEauthorblockN{Bartlomiej Sniezynski}
\IEEEauthorblockA{\textit{Institute of Computer Science} \\
\textit{AGH University of Science and Technology}\\
Krakow, Poland \\
bartlomiej.sniezynski@agh.edu.pl}
\and
 \IEEEauthorblockN{Michael Baron}
\IEEEauthorblockA{\textit{Department of Mathematics and Statistics} \\
\textit{American University}\\
Washington, DC, USA \\
baron@american.edu}
\and
 \IEEEauthorblockN{Nathalie Japkowicz}
 \IEEEauthorblockA{\textit{Department of Computer Science} \\
 \textit{American University}\\
 Washington, DC, USA \\
 japkowic@american.edu}
}

© 2022 IEEE.  Personal use of this material is permitted.  Permission from IEEE must be obtained for all other uses, in any current or future media, including reprinting/republishing this material for advertising or promotional purposes, creating new collective works, for resale or redistribution to servers or lists, or reuse of any copyrighted component of this work in other works.

DOI: \href{https://ieeexplore.ieee.org/document/9671962}{10.1109/BigData52589.2021.9671962}

\newpage
\maketitle

\begin{abstract}
Detecting relevant changes in dynamic time series data in a timely manner is crucially important for many data analysis tasks in real-world settings. Change point detection methods have the ability to discover changes in an unsupervised fashion, which represents a desirable property in the analysis of unbounded and unlabeled data streams.
However, one limitation of most of the existing approaches is represented by their limited ability to handle multivariate and high-dimensional data, which is frequently observed in modern applications such as traffic flow prediction, human activity recognition, and smart grids monitoring. 
In this paper, we attempt to fill this gap by proposing \WATCH{}, a novel Wasserstein distance-based change point detection approach that models an initial distribution and monitors its behavior while processing new data points, providing accurate and robust detection of change points in dynamic high-dimensional data.  
An extensive experimental evaluation involving a large number of benchmark datasets shows that \WATCH{} is capable of accurately identifying change points and outperforming state-of-the-art methods.
\end{abstract}


\section{Introduction}

Change point detection is 
of fundamental importance in modern data analysis problems characterized by dynamic time series data in which the statistical properties of the distribution change over time. One attractive characteristic of change point algorithms is their ability to analyze data in an unsupervised manner \cite{doi:10.1080/07474940008836437,aminikhanghahi2017survey}, which is ideal in analytical settings where data flows in an unbounded way and without annotations, as in data streams \cite{gama2010knowledge}. The change point detection problem is also implicitly connected to anomaly detection \cite{DS2021,chandola2009anomaly,faber2021autoencoder,ryan2019pattern} and concept drift detection in data streams \cite{krawczyk2018combining,souza2020unsupervised}. However, while change point detection focuses on the identification of points where the data distribution changes (and remains in that state for a certain amount of time), anomaly detection focuses on identifying out of distribution (even single) data points. Although a similarity between change point detection and concept drift detection can be noticed, change point detection assumes a more general connotation of change, whereas a more bounded characterization is made in concept drift detection \cite{webb2016characterizing,wares2019}.
 
An analysis of the existing literature in change point detection reveals a large number of approaches.
However, modern real-world applications such as traffic flow prediction \cite{essien2021deep}, human activity recognition \cite{reyes2016transition}, and renewable energy plants monitoring in smart grids \cite{ceci2015big,habault2019detecting,ceci2020echad} and settings characterized by geo-distributed data \cite{zhu2021high,BDR2019} are strongly characterized by multivariate and often high-dimensional data. 
One limitation of most change point detection methods 
described in the literature is that they perform well strictly with univariate data or multivariate data with a limited number of dimensions \cite{prabuchandran2021change}. As soon as the data dimensionality increases, their performance presents a drop in terms of accuracy or a significant increase in terms of execution time, which represents a known obstacle for online data analysis algorithms that rely on 
timely feedback from the change point detection procedure. 

In this paper, we address this gap by proposing \WATCH{}, a method based on the Wasserstein distance specifically targeted for high-dimensional time series data. The method models an initial distribution and monitors its behavior while processing new incoming data points. Change points are identified considering the Wasserstein distance between data points and the modeled distribution, which evolves over time. When a change point is detected, the previously known distribution is replaced to reflect the new distribution. 
An extensive experimental evaluation shows that our method significantly outperforms state-of-the-art change point detection methods on high-dimensional datasets. Moreover, the method achieves competitive performance on many of the 
benchmark datasets commonly used in the change point detection community. 

The remainder of the paper is structured as follows. Section \ref{sec:background} discusses an overview of existing change point detection approaches in the literature. Section \ref{sec:method} describes our proposed method. Section \ref{sec:results} provides details on our experimental setup, followed by Section \ref{sec:discussion} which discusses our results. Finally, Section \ref{sec:conclusion} concludes the paper, summarizing our contributions and showcasing directions for future work.



\section{Background}
\label{sec:background}

In this section we give an overview of some popular change point detection approaches in the literature. Methods can be broadly grouped in three main categories: offline, Bayesian, and non-parametric.
Offline approaches include likelihood ratio methods, which analyze the probability distributions of data before and after a candidate change point, in order to confirm the candidate as an actual change point if the two distributions are significantly different. One notable example is the CUSUM method, which triggers a detection when the cumulative sum of observations exceeds a given threshold value. A method known as At Most One Change \cite{hinkley1970inference} generalized this approach to testing for differences in the maximum likelihood. Typically, methods in this class calculate the logarithm of the likelihood ratios within two consecutive intervals and monitor it over time \cite{kawahara2012sequential,basseville1993detection}. Another variant of this method considers dimensionality reduction as a pre-processing step \cite{Plagianakos2015}. One major drawback of such methods is that they rely on predefined parametric models, which make them less flexible in real-world scenarios \cite{aminikhanghahi2017survey}.

Some offline methods can discover multiple change points within a time series, performing a greedy binary segmentation according to a cost function. Examples of methods in this class are Binary Segmentation \cite{scott1974cluster} and Segment Neighborhoods \cite{auger1989algorithms}.  An interesting variant of the binary segmentation approach which leverages the CUSUM statistic is Wild Binary Segmentation \cite{fryzlewicz2014wild}. Extensions leveraging pruning approaches and presenting an improved time efficiency include Pruned Exact Linear Time \cite{killick2012optimal} and Robust Functional Pruning Optimal Partitioning \cite{fearnhead2019changepoint}, which is also robust to outliers. It is noteworthy that these methods are restricted to the analysis of univariate datasets.

Probabilistic methods include offline and online Bayesian methods. A non-conventional offline change point detection approach is the Prophet method: a Bayesian time series forecasting approach that supports time series with change points \cite{taylor2018forecasting}. 
A popular online method in this class is Bayesian Online Change Point Detection \cite{adams2007bayesian} which is based on the assumption that a data sequence may be divided into non- overlapping states partitions and the data points in each state are independent and identically distributed according to some probability distribution. Methods in this class calculate the run length distribution according to Bayes' theorem and update the corresponding statistics. Change point prediction is then carried out by comparing probability values. Recent extensions include support for Model Selection \cite{knoblauch2018doubly} and spatio-temporal models \cite{knoblauch2018spatio}. Variants of Bayesian methods are based on the Gaussian Process (GP) method, which generalizes a Gaussian distribution as a collection of random variables, any finite number of which have a joint Gaussian distribution \cite{chandola2010scalable}, \cite{brahim2004gaussian}.

Non-parametric methods for change point detection are based on hypothesis testing, using a test statistic and a custom threshold. Examples of methods in this class include \cite{haynes2017computationally}, Kernel Change-Point Analysis \cite{harchaoui2009kernel}, and Energy Change Point \cite{matteson2014nonparametric}.
It is worthwhile mentioning that non-conventional change detection approaches based on deep neural networks are also investigated in the literature \cite{ceci2020echad,finder2021}, even though the scope of this study focuses on purely statistical change point detection.

\begin{table}
\centering
\caption{Change point detection methods featured in our experiments.}
\begin{tabular}{lll}
\textbf{Method} & \textbf{Name} & \textbf{Reference} \\
\hline
\textbf{Offline} & & \\
At Most One Change & AMOC & \cite{hinkley1970inference} \\
Binary Segmentation & BINSEG & \cite{scott1974cluster} \\
Segment Neighborhoods & SEGNEIGH & \cite{auger1989algorithms} \\
Wild Binary Segmentation & WBS & \cite{fryzlewicz2014wild} \\
Pruned Exact Linear Time & PELT & \cite{killick2012optimal} \\
Robust Functional Pruning Optimal Partitioning & RFPOP & \cite{fearnhead2019changepoint} \\
\hline
\textbf{Bayesian} & & \\
Bayesian Online Change Point Detection & BOCPD & \cite{adams2007bayesian} \\
BOCPD with Model Selection & BOCPDMS & \cite{knoblauch2018doubly} \\ 
BOCPD with Spatio-temporal models & RBOCPDMS & \cite{knoblauch2018spatio} \\
Prophet & PROPHET & \cite{taylor2018forecasting} \\
\hline
\textbf{Non-parametric} & & \\
Nonparametric Change Point Detection & CPNP & \cite{haynes2017computationally} \\
Kernel Change-Point Analysis & KCPA & \cite{harchaoui2009kernel} \\
Energy Change Point & ECP & \cite{matteson2014nonparametric} \\
\end{tabular}
\label{tbl:methods-framework}
\end{table}

\section{Method}
\label{sec:method}
In this section we describe \WATCH{}, our novel change point detection approach based on the Wasserstein distance. We divide the discussion in two stages: in the first subsection we introduce the Wasserstein distance discussing its potential for the task at hand; in the second subsection we provide algorithmic details about our method. 

\subsection{Wasserstein distance}

The Wasserstein metric is a distance function defined by a probability distribution on a given metric space $M$. 
An intuitive explanation of the metric provided in the literature views the distribution as a unit amount of earth (soil) piled on $M$, where the metric represents the minimum "cost" of turning one pile into the other. This cost is assumed to be the amount of earth that needs to be moved, multiplied by the mean distance it has to be moved. Because of this analogy, the metric is known in computer science as the earth mover's distance. Recent works in the machine learning literature incorporated this distance in model pipelines and showed its potential in providing disentangled representations \cite{gaujac2021learning}.

According to the Wasserstein distance definition in \cite{hallin2021multivariate}, let $P(\mathbb{R}^d)$ be the set of Borel probability measures on $\mathbb{R}^d$ and $P_p(\mathbb{R}^d)$ be the subset of such measures with a finite moment of order $p \in [1, \inf)$. For P, Q $\in P(\mathbb{R}^d)$, let $\Gamma(\text{P},\text{Q})$ be the set of probability measures $\gamma$ on $\mathbb{R}^d \times \mathbb{R}^d$ with marginals P and Q, i.e., such that $\gamma(B \times \mathbb{R}^d) = \text{P}(B)$ and $\gamma(\mathbb{R}^d \times B) = \text{Q}(B)$ for Borel sets $B \subseteq \mathbb{R}^d$. $dist(\dot)$ is the Euclidean norm. The p-Wasserstein distance between P, Q $ \in P_p(\mathbb{R}^d)$ is:
\begin{equation}
    W_p(\text{P}, \text{Q}) = {(\inf_{\gamma \in \Gamma(P,Q)} \int_{\mathbb{R}^d \times \mathbb{R}^d} dist(x - y)^p \ \partial \gamma(x,y))}^{\frac{1}{p}}.
\end{equation}
In terms of random variables $X$ and $Y$ with laws P and Q, respectively, the p-Wasserstein distance is the smallest value of $\{{E(dist(X - Y))^p}\}^\frac{1}{P}$ over all possible joint distributions $\gamma \in \Gamma(\text{P}, \text{Q})$ of (X, Y).
It is important to note that the original formulation of the Wasserstein distance is intractable in nature. Therefore, practitioners estimate it relying on an approximated implementation. 
Intuitively, given a $k$-dimensional vector space $V \subset \mathbb{R}^k$, we can map the distributions $(P,Q)$ to  corresponding sets of $k$-dimensional data points $G,H \subseteq V$. In our work, the Wasserstein distance $W_P$ is calculated using the SAG approximated computational method as explained in \cite{hallin2021multivariate}. From now on, we refer to the approximated version of the Wasserstein distance as $W_A \approx W_p$.

\begin{figure}
    \centering
    \includegraphics[scale=0.3]{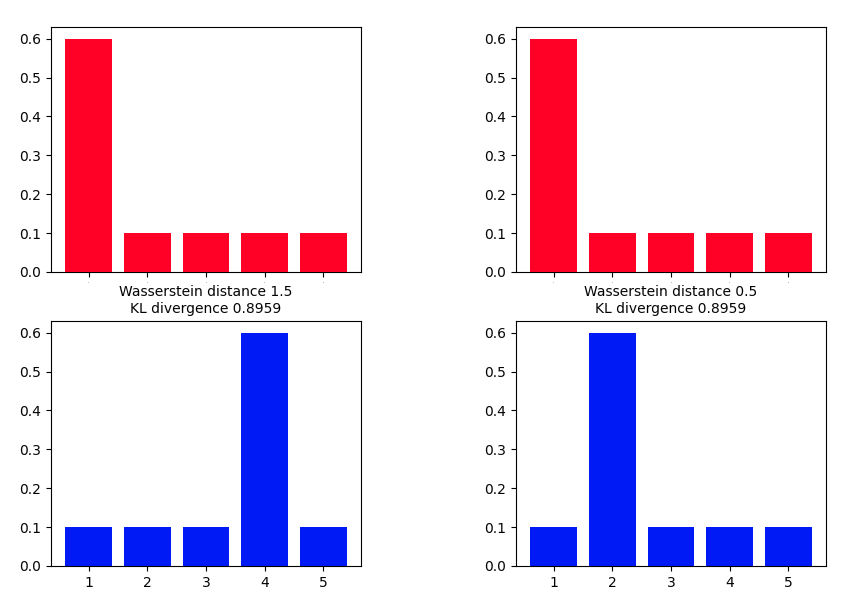}
    \caption{Difference between Wasserstein distance and Kullbach-Leibler (KL) divergence.}
\label{fig:wasserstein-dist}
\end{figure}

The rationale for the adoption of the Wasserstein distance in our approach is that, unlike the Kullback-Leibler divergence, it is a probability metric that measures the work required to transport the probability mass from one distribution state to another. This characteristic provides a smooth and significant representation of the distance between distributions, making it a good candidate in domains where analyzing the similarity of outcomes is more relevant than an exact likelihood matching. 
Figure \ref{fig:wasserstein-dist} shows an intuitive representation of meaningful information carried by the Wasserstein distance: although the measures between the red and the blue distributions are the same in terms of Kullback-Leibler divergence, the Wasserstein distance measures ``the work" required to transport the probability mass from the red state to the blue state.

It is noteworthy that the Wasserstein distance is gaining traction in recent machine learning methods and notably in neural network architectures such as Wasserstein Auto-Encoders (WAE), that share many of the properties of Variational Auto-Encoders (VAEs) such as stable training, encoder-decoder architecture, and a nice latent manifold structure, while also generating samples of higher quality \cite{tolstikhin2018wasserstein}.
As indicated in \cite{gaujac2021learning}, a WAE encodes the data into a low-dimensional latent space similarly to a VAE. However, instead of regularizing each data point, it regularizes the averaged encoding distribution. This approach allows the encoding distribution to capture essential information and still match the prior distribution.

Our work builds upon distance-based change point detection approaches, which have shown success in the previous literature \cite{dasu2006,Aminikhanghahi2019RealTimeCP}. Our intuition is that adopting a more informative distance metric such as the Wasserstein distance, which benefits from the properties described above, could trigger more timely and accurate detection of change points. This assumption may be particularly relevant in high-dimensional datasets, where the large number of dimensions adds complexity to the change point detection task, causing phenomena such as collinearity, feature noise, and fractional distance when few of the features values change.

\subsection{Algorithm}

The input of the algorithm is a time series of data samples $\mathbf{I} = (I_1, I_2, \ldots, I_T)$, where each data sample is a $d$-dimensional vector: $I_t\in \mathbb{R}^d$. The parameters of the algorithm are the following:
\begin{itemize}
    \item \textit{$\kappa$} -- the minimum number of points that must be present in the current representation of distribution to trigger change point detection. This parameter controls how many samples are required before knowing the current distribution well enough for change point detection.
    \item \textit{$\mu$} -- the maximum number of points that can be present in the current representation of distribution. This parameter allows to control the size of the memory preventing possible bottlenecks deriving from a large number of samples.
    \item \textit{$\epsilon$} -- the ratio controlling how distant the samples can be from the current distribution to be still considered a part of the same.
    \item \textit{$\omega$} -- size of the mini-batch in which the data are processed. 
\end{itemize}

Intuitively, our method learns a new representation of a distribution $D$ in an unsupervised manner using an initial set of data points. Subsequently, the method incrementally stores newly available samples in a buffer. Once the desired capacity of the buffer (which is controlled by the $\kappa$ parameter) is reached, the change detection procedure is activated. In this process, we collect newly arriving points in mini-batches  of length $\omega$ (sliding non-overlapping windows), and calculate the Wasserstein distance of every mini-batch $B$ with respect to $D$.
If the distance is lower than the desired threshold $\gamma$ (controlled by the $\epsilon$ parameter), we qualify $B$ as belonging to $D$. The threshold is determined as:

\begin{equation}
    \eta(D, \ \epsilon) = \ \epsilon \ \max_{B \in D}  \ W_A(B, D) .
\label{eq:threshold}
\end{equation}

Consequently, $D$ is updated to reflect the new data points. On the contrary, if the distance is higher than $\gamma$, we assume that the current mini-batch $B$ being analyzed belongs to a different distribution. Therefore, we annotate this point in time as a change point and start creating a new distribution to replace $D$.  Once $D$ reaches its storage capacity controlled by $\mu$, the oldest samples are removed to accommodate new ones, in order to keep an up-to-date representation of the distribution $D$.
A pseudo-code of the algorithm is presented in Algorithm \ref{alg:main}.
A significant advantage of our approach is that it compares two sets of points instead of working on the parameters of specific distributions. This mechanism allows to compare distributions of unknown types, without limiting the scope to only the most popular ones, such as Normal distributions.
In the following sections we describe our experimental setup and discuss our results.

\begin{algorithm}
\SetAlgoLined
\LinesNumbered
\SetKwInOut{Input}{Input}
\SetKwInOut{Parameter}{Parameters}
\Input{$\mathbf{I} = (I_1, I_2, \ldots, I_T)$ Time series data}
\medskip 
\KwResult{$R$: Set of change points}
\medskip 
\Parameter{$\kappa$ \ Min points required in current distrib. before starting detection \\
$\mu$ Max points stored for current distrib. \\
$\epsilon$ \ Threshold ratio for the distance value \\
$\omega$ \ Mini-batch size}
\medskip 
Set current distribution $D \gets \emptyset$ \\ 
$R \gets \emptyset$ \\
\medskip 
$\mathbf{B} \gets $ Process $\mathbf{I}$ into sequence $(B_1, B_2, \ldots, B_{\frac{T}{\omega}})$ of non-overlapping mini-batches $B_i$ of size $\omega$ \\
\For{$B_i \in \mathbf{B}$}{
    \eIf{$|D| < \kappa $}{
        $D \gets D \cup B_i$ \\
        \If{$|D| >= \kappa$}{
            Determine $\eta$ from Eq. (\ref{eq:threshold}) with $D$ and $\epsilon$ \\
        }
    }{
        $\upsilon \gets W_A(B_i, D)$ \\
        \eIf{$\upsilon > \eta$}{
            Set change point $c \gets i * \omega$ \\
            $R \gets R \cup \{c\}$ \\
            $D \gets \{B_i\}$ \\
        }{
            \If{$|D| < \mu$}{
                $D \gets D \cup B_i$ \\
                Determine $\eta$ from Eq. \ref{eq:threshold} with $D$ and $\epsilon$ \\
            }
        }
        
    }
} 
\caption{Pseudo-code of the proposed \WATCH{} method}
\label{alg:main}
\end{algorithm}

\section{Experimental setup}
\label{sec:results}


In this study we adopt the Turing Change Point Detection (TCPD) benchmark \cite{burg2020evaluation} in order to compare the results obtained with our method with state-of-the-art algorithms. In the following subsection, we introduce the benchmark and motivate its adoption. Then, we discuss our additional datasets and the metrics used for the evaluation of results. 

\subsection{Turing Change Point Detection (TCPD) benchmark}
\label{subsec:benchmark}

This benchmark allows us to evaluate the 13 algorithms described in Table \ref{tbl:methods-framework} on 42 time series datasets. There is also the possibility to extend the benchmark to new methods and datasets, which provides a solid ground for the evaluation of new change point detection algorithms designed by researchers and practitioners. It is also noteworthy that many existing algorithms work just on univariate data and cannot handle multivariate datasets. Moreover, the majority of the datasets (38 out of 42) available in the benchmark are univariate (or  single-dimensional), and only 4 are multivariate (or multi-dimensional). Furthermore, these 4 datasets (\textit{apple, bee\_waggle\_6, occupancy, run\_log}) present a number of dimensions ranging from 2 to 4. For this reason, we extended the benchmark with 5 datasets with a significantly higher number of dimensions, ranging from 17 to 784. Further details of the added datasets are reported in Section \ref{subsec:datasets}. Table \ref{tab:datasets} presents all datasets used in our experimental evaluation.

There are two modes of experiments in the TCPD benchmark:
\begin{itemize}
    \item \textit{default} -- in this mode the algorithms are run with default parameters;
    \item \textit{best} -- in this mode the algorithms are run with various combinations of parameter values (grid search) and the best configuration is selected.
\end{itemize}
Summary results obtained with the \textit{default} and \textit{best} mode across all datasets are reported in Table \ref{tab:summary_results}, whereas detailed results obtained with the \textit{best} mode on each single dataset are reported in Table \ref{tab:f1_multi},  \ref{tab:cover_multi}, \ref{tab:f1_uni}, and \ref{tab:cover_uni}. 

\begin{table}
\centering
\caption{Multivariate datasets analyzed}
\begin{tabular}{l|c|c}
Name            &  Samples      & Dimensions \\
\midrule
apple           & 622           & 2         \\
bee\_waggle\_6  & 609           & 4         \\
gas\_sensor     & 1000          & 128       \\
har             & 1623          & 561        \\
mnist           & 131           & 784        \\
movement        & 360           & 90        \\
occupancy       & 509           & 4         \\
run\_log        & 376           & 2         \\
traffic         & 135           & 17        \\
\end{tabular}
\label{tab:datasets}
\end{table}

\subsection{Additional high-dimensional datasets}
\label{subsec:datasets}
We added the following datasets to evaluate the performance of the methods on more challenging data presenting a higher number dimensions than 4: 
\begin{enumerate}
    \item \textbf{har:} The Human Activity Recognition dataset introduced in \cite{Anguita2013APD} contains records of 6 types of human activities including w alking, sitting, standing. Given the sequential nature of the data, it is straightforward to determine change points based on activity changes in the labels for each human participant. We perform experiments separately for all human participants and report the average values of the metrics. 
    \item \textbf{gas\_sensor:} The Gas Sensor Array Drift Dataset \cite{misc_gas_sensor_array_drift_dataset_at_different_concentrations_270} contains measurements from 16 sensors exposed to 6 different gases at various concentration levels. Given the large number of dimensions (128) and time points (13910), we selected a portion of 1000 samples in order to allow all methods to complete their computation in a feasible amount of time. Change points are determined in the same way as in the \textbf{har} dataset, taking into account class changes in the time series data.
    \item \textbf{movement:} The Libras movement dataset contains records of hand movements types in official Brazilian signal language \cite{misc_libras_movement_181}. There are 360 samples described by 90 dimensions. We determine the change points based on labels' changes in the same way as in the \textbf{har} dataset.
    \item \textbf{mnist:} The MNIST dataset \cite{mnist} contains images of handwritten digits. We reproduced a sequential scenario selecting 131 images, and a varying number of images (from 6 to 35) for each class to obtain a more challenging scenario with non-evenly spaced change points. Each image has a size of $28 \times 28$, which is flattened to obtain a 784-dimensional feature vector.
    \item \textbf{traffic:} This dataset contains urban traffic records of the city of Sao Paulo in Brazil \cite{misc_behavior_of_the_urban_traffic_of_the_city_of_sao_paulo_in_brazil_483}. There are 135 samples described by 17 dimensions. We first perform discretization on the \textit{slowness} attribute (continuous) to obtain a discrete attribute with three labels (\textit{low}, \textit{medium}, and \textit{high}). Then, we determine change points based on label changes in the sequential data. 
\end{enumerate}

\subsection{Metrics}
In our experiments, we adopt the metrics provided by the authors of the TCPD benchmark \cite{burg2020evaluation} to keep consistency and to allow an easy comparison for future works. As indicated in \cite{burg2020evaluation}, the existing metrics for change point detection can be divided between classification and clustering metrics. One metric for each category is adopted: F1-score for classification and Covering for clustering.  

\subsubsection*{F1-score}
Change point detection algorithms may be evaluated in terms of classification between change-point and non-change point classes \cite{burg2020evaluation}. Usually, datasets are affected by class imbalance due to a low number of change points. Therefore, metrics that are highly vulnerable to class imbalance are not suitable for this case. Examples of more robust metrics are:
\begin{itemize}
    \item $P$ -- \textit{Precision}: the ratio of correctly detected change points over the number of all detected change points,
    \item $R$ -- \textit{Recall}: the ratio of correctly detected change points over the number of actual change points.
\end{itemize}

The \textit{F1-score} can be applied to provide a single performance measure, incorporating both Precision and Recall. The F1 score is defined as:

\begin{equation}
    F_1 = \frac{P  R}{P + R}.
\end{equation}
As the TCPD benchmark may contain many ground truth sets of change points provided by human annotators, Precision and Recall are adjusted to handle this situation. This aspect implies that only detected change points that are not present in any ground truth are considered false positives, while the Recall is calculated as the average for each source of ground truth to encourage finding all change points from all ground truths (from all annotators). The details on those metrics are presented in \cite{burg2020evaluation} and \cite{Martin2004}. The TCPD benchmark also considers the margin of error around the actual change points to allow for minor discrepancies. In the results provided by the original paper and in our work, we use a margin of error of value 5.

\vspace{2 pt}
\subsubsection*{Covering metric (Cover)} 
The covering metric was initially presented in \cite{Arbelaez2011} to assess the quality of the segmentation for multiple segments (subsets of pixels representing separate objects) in the image domain. Simply put, the covering metric describes how well the predicted segments cover ground truth segments. In our case, each segment corresponds to a specific subsequence of contiguous points from the input time series. In order to define the covering metric, we first need to introduce the Jaccard Index. It is a statistic used for measuring the similarity of sample sets. It is also known as Intersection over Union, as it measures the overlap between the predicted segmentation and the ground truth. The Jaccard Index is defined as: 

\begin{equation}
    J(A, A') = \frac{|A \cap A'|}{|A \cup A'|}
\end{equation}

The definition of the covering metric of partition $G$ by partition $G'$ is introduced in \cite{Arbelaez2011} for images segmentation and adjusted to time series data in \cite{burg2020evaluation}:
\begin{equation}
    Cover(G',G) = \frac{1}{T} \sum_{A \in G} |A|  \max_{A' \in G'} J(A, A')
\end{equation}

To determine a single measure of performance while having a collection of ground truth partitions provided by the human annotators and a partition given by an algorithm, the average of $Cover(G', G)$ for all annotators is computed.

\subsection{Implementation}
\WATCH{} is implemented in Python v3.9 and numpy v1.19.5. We also used the R implementation of the Wasserstein distance provided in \cite{hallin2021multivariate}, and called it from the Python code using the rpy2 v3.4.5 bridge. All experiments are run on a machine with an Intel Core i7-8750H CPU, GeForce GTX 1050 Ti Mobile GPU and 32 GB of RAM.

\section{Results and discussion}
\label{sec:discussion}

In the following, we discuss our experimental results in three subsections: \textit{A)} summary results across all datasets considering \textit{default} and \textit{best} modes, \textit{B)} detailed results on multivariate (multi-dimensional) datasets, and \textit{C)} detailed results on univariate (uni-dimensional) datasets.

\subsection{Summary results analysis}

Table \ref{tab:summary_results} shows the summary of the average results across all datasets and methods. The layout of the table organizes results into four sections based on dimensions of the data (univariate, multivariate), and the mode of the experiments (\textit{default, best)}. The values correspond to the average of runs for each section.

\begin{table}[]
    \scriptsize
    \centering
        \caption{Results organized into sections based on dimensions of the data and the mode of experiments. The values correspond to the average of runs for each section. Blank values indicate that the method is not suited for multidimensional data. The highest values for each section of the table are shown in bold.}
    \scalebox{0.9}{
\begin{tabular}{lrr|rrrr|rr}
 & \multicolumn{4}{c}{Univariate} & \multicolumn{4}{c}{Multivariate} \\\cmidrule(lr){2-5}\cmidrule(lr){6-9}
 & \multicolumn{2}{c}{Default} & \multicolumn{2}{c}{Best} & \multicolumn{2}{c}{Default} & \multicolumn{2}{c}{Best} \\\cmidrule(lr){2-3}\cmidrule(lr){4-5}\cmidrule(lr){6-7}\cmidrule(lr){8-9}
Algorithm & Cover & F1 & Cover & F1 & Cover & F1 & Cover & F1\\\cmidrule(r){1-1}\cmidrule(lr){2-5}\cmidrule(l){6-9}
\textsc{amoc}     & 0.680 & 0.677 & 0.726 & 0.778 &       &       &       &       \\
\textsc{binseg}   & \textbf{0.682} & \textbf{0.716} & 0.760 & 0.833 &       &       &       &       \\
\textsc{bocpd}    & 0.614 & 0.675 & 0.769 & 0.857 & \textbf{0.486} & \textbf{0.558} & \textbf{0.717} & 0.798 \\
\textsc{bocpdms}  & 0.626 & 0.507 & 0.737 & 0.620 & 0.249 & 0.235 & 0.317 & 0.292 \\
\textsc{cpnp}     & 0.514 & 0.594 & 0.538 & 0.648 &       &       &       &       \\
\textsc{ecp}      & 0.505 & 0.592 & 0.711 & 0.789 & 0.456 & 0.496 & 0.634 & 0.683 \\
\textsc{kcpa}     & 0.068 & 0.121 & 0.619 & 0.679 & 0.162 & 0.187 & 0.565 & 0.662 \\
\textsc{pelt}     & 0.666 & 0.692 & 0.706 & 0.766 &       &       &       &       \\
\textsc{prophet}  & 0.542 & 0.495 & 0.574 & 0.537 &       &       &       &       \\
\textsc{rbocpdms} & 0.579 & 0.411 & 0.422 & 0.349 & 0.194 & 0.191 & 0.077 & 0.102 \\
\textsc{rfpop}    & 0.374 & 0.485 & 0.403 & 0.517 &       &       &       &       \\
\textsc{segneigh} & 0.658 & 0.659 & 0.763 & 0.832 &       &       &       &       \\
\textsc{wbs}      & 0.319 & 0.408 & 0.417 & 0.519 &       &       &       &       \\
\textsc{zero}     & 0.573 & 0.658 & 0.573 & 0.658 & 0.255 & 0.325 & 0.255 & 0.325 \\
\cmidrule(r){1-1}\cmidrule(lr){2-5}\cmidrule(l){6-9}

\textsc{watch}   & 0.540 & 0.570 & \textbf{0.797} & \textbf{0.898} & 0.427 & 0.488 & 0.713 & \textbf{0.836} \\
\cmidrule(r){1-1}\cmidrule(lr){2-5}\cmidrule(l){6-9}
\end{tabular}

    }
    \label{tab:summary_results}
\end{table}

Considering the results obtained for the \textit{best} mode with multivariate datasets, the \WATCH{} method outperforms all other ones in terms of F1. The second best performing method, BOCPD, has an average F1 of 0.798, while \WATCH{} has an average F1 of 0.836, resulting in a 4.76\% improvement. The other methods, ECP and KCPA, are strongly outperformed by 22.40\% and 26.28\%, respectively. Focusing on the Cover metric, the \WATCH{} method performs very similarly to the best method BOCPD, reporting a sligthly lower value (0.55\%). It is noteworthy, however, that \WATCH{} achieves significantly better Cover than the ECP and KCPA. 
Focusing on the results obtained for univariate datasets, it is noteworthy that the proposed method \WATCH{} outperforms other methods in terms of both Cover and F1 also for these simpler datasets. 


\subsection{Results on multivariate datasets}

\begin{table*}[h]
    \centering
        \caption{F1 for multivariate datasets and methods. Values are unavailable either due to a failure (F) or the method timing out (T). Highest values for each time series are shown in bold.}
    \begin{tabular}{lcccccc|c}
Dataset & \textsc{bocpd} & \textsc{bocpdms} & \textsc{ecp} & \textsc{kcpa} & \textsc{rbocpdms} & \textsc{zero} & \textsc{watch}\\
\hline
\verb+apple+ & 0.916 & 0.445 & 0.745 & 0.634 & F/T & 0.594 & \textbf{0.941}\\
\verb+bee_waggle_6+ & \textbf{0.929} & 0.481 & 0.233 & 0.634 & 0.245 & \textbf{0.929} & \textbf{0.929}\\
\verb+gas_sensor+ & 0.619 & T & \textbf{0.634} & 0.444 & F/T & 0.091 & 0.604\\
\verb+har+ & 0.792 & F/T & 0.885 & 0.831 & F/T & 0.148 & \textbf{0.908}\\
\verb+mnist+ & 0.533 & T & 0.625 & 0.706 & F/T & 0.167 & \textbf{0.900}\\
\verb+movement+ & 0.769 & T & 0.651 & 0.571 & T & 0.125 & \textbf{0.833}\\
\verb+occupancy+ & 0.919 & 0.735 & 0.932 & 0.812 & F/T & 0.341 & \textbf{0.968}\\
\verb+run_log+ & \textbf{1.000} & 0.469 & 0.990 & 0.909 & 0.380 & 0.446 & 0.671\\
\verb+traffic+ & 0.703 & 0.500 & 0.451 & 0.414 & 0.296 & 0.083 & \textbf{0.773}\\
\hline
\end{tabular}

    \label{tab:f1_multi}
\end{table*}

\begin{table*}[h]
    \centering
        \caption{Cover for multivariate datasets and methods. Values are unavailable either due to a failure (F) or the method timing out (T). Highest values for each time series are shown in bold.}
    \begin{tabular}{lcccccc|c}
Dataset & \textsc{bocpd} & \textsc{bocpdms} & \textsc{ecp} & \textsc{kcpa} & \textsc{rbocpdms} & \textsc{zero} & \textsc{watch}\\
\hline
\verb+apple+ & \textbf{0.846} & 0.720 & 0.758 & 0.462 & F/T & 0.425 & 0.748\\
\verb+bee_waggle_6+ & \textbf{0.891} & 0.889 & 0.116 & 0.730 & 0.205 & \textbf{0.891} & \textbf{0.891}\\
\verb+gas_sensor+ & 0.742 & T & \textbf{0.781} & 0.435 & F/T & 0.094 & 0.743\\
\verb+har+ & 0.732 & F/T & \textbf{0.832} & 0.729 & F/T & 0.084 & 0.802\\
\verb+mnist+ & 0.464 & T & 0.611 & \textbf{0.698} & F/T & 0.123 & 0.697\\
\verb+movement+ & \textbf{0.766} & T & 0.732 & 0.384 & T & 0.067 & 0.699\\
\verb+occupancy+ & 0.645 & 0.556 & 0.666 & 0.581 & F/T & 0.236 & \textbf{0.671}\\
\verb+run_log+ & \textbf{0.824} & 0.327 & 0.819 & 0.729 & 0.329 & 0.304 & 0.603\\
\verb+traffic+ & 0.546 & 0.357 & 0.387 & 0.332 & 0.158 & 0.073 & \textbf{0.561}\\
\hline
\end{tabular}

    \label{tab:cover_multi}
\end{table*}

Results in terms of the F1 and Cover metrics values for multivariate datasets are reported in Table \ref{tab:f1_multi} and \ref{tab:cover_multi}. In this work we are particularly focused on high-dimensional datasets since they most closely represent complex real-world scenarios that are not supported by the majority of change point detection methods (we note that only 6 out of 14 competitor methods considered in our experiments support the analysis of multivariate data). In the following, we briefly discuss results obtained on each multivariate dataset, whereas results on univariate datasets are reported separately in the following subsection. Results obtained by our method in terms of both F1 and Cover metrics follow a similar pattern. For the sake of conciseness, we focus our discussion on the F1 metric, which we believe is the most representative in most scenarios with time series data: 

\begin{figure*}[]
    \centering
    \includegraphics[width=0.9\textwidth]{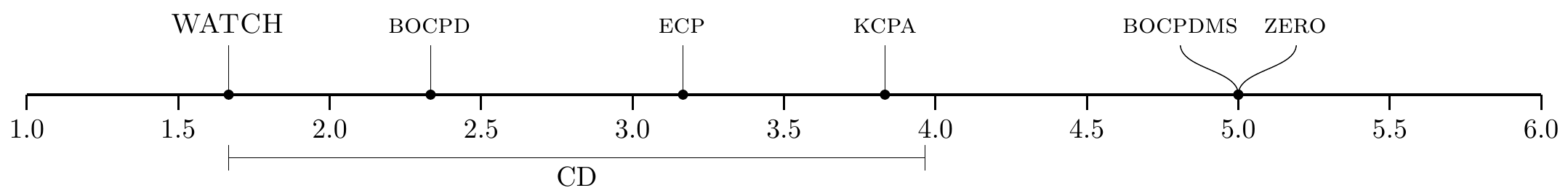}
    \caption{F1 rank plot for multivariate data in the \textit{best} mode.}
\label{fig:rankplot_f1}
\end{figure*}

\begin{figure*}[]
    \centering
    \includegraphics[width=0.9\textwidth]{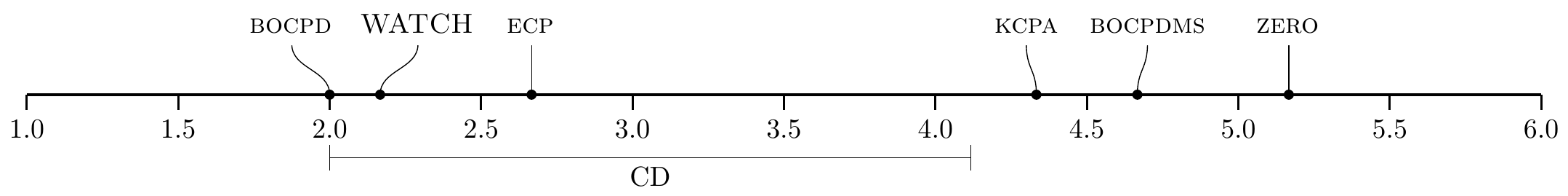}
    \caption{Cover rank plot for multivariate data in the \textit{best} mode.}
\label{fig:rankplot_cover}
\end{figure*}

\begin{itemize}
    \item \textbf{apple} --  \WATCH{} outperforms all other competitors, improving the results by at least 2.72\% considering the second best performing method, BOCPD.
    \item \textbf{bee\_waggle\_6} --  \WATCH{} achieves the same performance as BOCPD and ZERO methods both in terms of F1 and Cover.
    \item \textbf{gas\_sensor} -- \WATCH{} achieves a lower F1 performance than ECP and BOCPD. Considering that the \WATCH{} behavior internally depends on a distance metric, we attribute this loss to a possible fractional distance issue depending on the large number of dimensions of this dataset (128). More specifically, if there are just few dimensions determining the change points, they may be harder to detect since the overall distance may not change significantly. However, it is noteworthy the performance in terms of F1 and Cover is still relatively close to the best performing methods: a 4.73\% margin in terms of F1 with respect to ECP, which loses with most of the other datasets analyzed. 
    \item \textbf{har} -- \WATCH{} significantly outperforms all other methods. Considering ECP, the second best performing method, \WATCH{} improves its F1 score by 2.59\%, and BOCPD by 14.64\%. We consider this result as a representative case of success, considering both the large number of dimensions in this dataset (561) and the significant real-world implications, including human activity recognition and fall detection among others. 
    \item \textbf{movement} -- \WATCH{} greatly outperforms other methods, with a 8.32\% improvement in F1 score over BOCPD. Pairwise comparisons with other methods reveal even greater improvements.
    \item \textbf{mnist} -- \WATCH{} presents a huge margin of improvement when compared to other methods: at least 27.47\% when compared to KCPA, the second best performing method. 
    \item \textbf{occupancy} -- \WATCH{} outperforms ECP, the second best performing method, by 3.86\% in terms of F1 score, achieving an almost perfect score (close to 1).
    \item \textbf{run logs} - \WATCH{} does not obtain competitive results on this dataset,  presenting significantly worse results than the other methods. However, it is important to note that this dataset has only 2 dimensions. A quick inspection at the data shows that one dimension (overall \textit{travel distance} of the runner) is monotonically increasing since it is accumulated over time, and therefore does not follow a typical data distribution. This data characteristic suggests that \WATCH{} is unable to correctly detect changes due to the Wasserstein distance values being dominated by one of the two dimensions, yielding a large number of false positives. We argue that this dataset is atypical in nature, and is not representative of our main target, i.e. high-dimensional data. In this special case, the outcome may be completely different by just excluding the cumulative distance as a dimension, and limiting the analysis of the other dimension (\textit{running pace}).  
    \item \textbf{traffic} - \WATCH{} significantly outperforms other methods, achieving an F1 score that is higher by $9.95\%$ than the second-best performing method, BOCPD, and by $54.6\%$ than the third-best performing method, BOCPDMS. We argue that this dataset is representative of our scope of analysis due to the strong real-world implication of traffic monitoring. 
\end{itemize}

As we can see, the \WATCH{} method achieves the best results with 7 out of 9 multivariate datasets. Most importantly, it provides consistent and robust detections with the datasets presenting the highest dimensionality (HAR and MNIST). 

In order to assess the statistical significance of our results, we adopt a rank plot for both metrics (F1 and Cover) computed by averaging the ranks obtained by the algorithm over all datasets  \cite{demvsar2006statistical}. More specifically, Holm's procedure was adopted to calculate the significant differences considering all the different algorithms \cite{holm1979simple}. Figure \ref{fig:rankplot_f1} shows that \WATCH{} strongly outperforms other methods in terms of F1. Looking at the figure, the pairwise difference between any two combinations of algorithms is significant if their distance exceeds the critical difference (CD). In terms of Cover, looking at Figure \ref{fig:rankplot_cover} reveals a slightly lower performance for \WATCH{}, which is not significantly worse than BOCPD. Moreover, it is noteworthy that \WATCH{} still obtains the best rank values compared to all other algorithms considered in our study.

\subsection{Univariate results analysis}
Although this work focuses on multivariate data, it is noteworthy that \WATCH{} presents a consistent performance also with the univariate datasets available in TCPD benchmark. Table \ref{tab:f1_uni} and Table \ref{tab:cover_uni} show the results in terms of F1 and Cover for all univariate datasets and algorithms. Overall, \WATCH{} achieves the best F1 results with 23 out of 38 datasets. Considering the Cover metric,  \WATCH{} shows the highest value with 20 datasets. Moreover, Table \ref{tab:summary_results}, shows that \WATCH{} achieves the best average values of F1 and Cover for univariate datasets in the \textit{best} experiment mode.

\begin{table*}[]
    \setlength{\tabcolsep}{4pt}
    \scriptsize
    \centering
        \caption{F1 for univariate datasets  (\textit{best} mode). Values are unavailable either due to a failure (F), the method timing out (T) or missing
values in the series unsupported by the method (M). Highest values for each time series are shown in bold.}
    \scalebox{0.96}{
    \begin{tabular}{lcccccccccccccc|c}
Dataset & \textsc{amoc} & \textsc{binseg} & \textsc{bocpd} & \textsc{bocpdms} & \textsc{cpnp} & \textsc{ecp} & \textsc{kcpa} & \textsc{pelt} & \textsc{prophet} & \textsc{rbocpdms} & \textsc{rfpop} & \textsc{segneigh} & \textsc{wbs} & \textsc{zero} & \textsc{watch}\\
\hline
\verb+bank+ & \textbf{1.000} & \textbf{1.000} & \textbf{1.000} & 0.500 & 0.054 & 0.182 & 0.333 & 0.400 & \textbf{1.000} & T & 0.015 & \textbf{1.000} & 0.043 & \textbf{1.000} & \textbf{1.000}\\
\verb+bitcoin+ & 0.507 & 0.690 & 0.733 & 0.533 & 0.611 & 0.648 & 0.665 & 0.735 & 0.446 & T & 0.284 & 0.735 & 0.690 & 0.450 & \textbf{0.806}\\
\verb+brent_spot+ & 0.465 & 0.670 & 0.609 & 0.239 & 0.607 & 0.641 & 0.553 & 0.586 & 0.249 & T & 0.521 & 0.586 & 0.564 & 0.315 & \textbf{0.677}\\
\verb+businv+ & \textbf{0.588} & \textbf{0.588} & \textbf{0.588} & 0.455 & 0.386 & 0.370 & 0.294 & 0.490 & 0.275 & 0.370 & 0.261 & \textbf{0.588} & 0.289 & \textbf{0.588} & \textbf{0.588}\\
\verb+centralia+ & 0.909 & \textbf{1.000} & \textbf{1.000} & \textbf{1.000} & \textbf{1.000} & 0.909 & \textbf{1.000} & \textbf{1.000} & 0.763 & 0.846 & \textbf{1.000} & \textbf{1.000} & 0.556 & 0.763 & 0.974\\
\verb+children_per_woman+ & 0.678 & 0.663 & 0.712 & 0.405 & 0.344 & 0.551 & 0.525 & 0.637 & 0.310 & 0.504 & 0.246 & 0.637 & 0.500 & 0.507 & \textbf{0.844}\\
\verb+co2_canada+ & 0.544 & 0.856 & \textbf{0.924} & 0.479 & 0.642 & 0.875 & 0.867 & 0.670 & 0.482 & 0.542 & 0.569 & 0.872 & 0.681 & 0.361 & 0.893\\
\verb+construction+ & 0.696 & 0.709 & 0.709 & 0.410 & 0.602 & 0.709 & 0.634 & 0.709 & 0.324 & 0.340 & 0.185 & 0.709 & 0.523 & 0.696 & \textbf{0.966}\\
\verb+debt_ireland+ & 0.760 & \textbf{1.000} & \textbf{1.000} & 0.892 & 0.958 & 0.980 & \textbf{1.000} & \textbf{1.000} & 0.469 & 0.748 & 0.824 & \textbf{1.000} & 0.538 & 0.469 & 0.958\\
\verb+gdp_argentina+ & 0.889 & \textbf{0.947} & \textbf{0.947} & 0.583 & 0.818 & 0.824 & 0.800 & \textbf{0.947} & 0.615 & 0.452 & 0.615 & \textbf{0.947} & 0.421 & 0.824 & \textbf{0.947}\\
\verb+gdp_croatia+ & \textbf{1.000} & 0.824 & \textbf{1.000} & 0.583 & \textbf{1.000} & 0.824 & 0.583 & 0.824 & 0.824 & 0.824 & 0.400 & 0.824 & 0.167 & 0.824 & \textbf{1.000}\\
\verb+gdp_iran+ & 0.696 & 0.652 & 0.862 & 0.492 & 0.620 & 0.824 & 0.734 & 0.808 & 0.652 & 0.737 & 0.636 & 0.808 & 0.576 & 0.652 & \textbf{0.889}\\
\verb+gdp_japan+ & \textbf{1.000} & 0.889 & \textbf{1.000} & 0.615 & 0.667 & \textbf{1.000} & 0.500 & 0.889 & 0.889 & 0.889 & 0.222 & 0.889 & 0.222 & 0.889 & \textbf{1.000}\\
\verb+global_co2+ & 0.929 & 0.929 & 0.889 & 0.458 & 0.667 & \textbf{0.929} & 0.667 & 0.929 & 0.463 & 0.547 & 0.293 & 0.929 & 0.250 & 0.846 & 0.846\\
\verb+homeruns+ & 0.812 & 0.829 & 0.829 & 0.650 & 0.650 & 0.812 & 0.829 & 0.812 & 0.723 & 0.397 & 0.661 & 0.812 & 0.664 & 0.659 & \textbf{0.974}\\
\verb+iceland_tourism+ & 0.947 & 0.947 & 0.947 & 0.486 & 0.391 & \textbf{1.000} & 0.486 & 0.643 & 0.220 & 0.667 & 0.200 & 0.947 & 0.200 & 0.947 & \textbf{1.000}\\
\verb+jfk_passengers+ & \textbf{0.776} & \textbf{0.776} & \textbf{0.776} & 0.650 & 0.602 & 0.651 & 0.437 & \textbf{0.776} & 0.354 & T & 0.491 & \textbf{0.776} & 0.437 & 0.723 & 0.723\\
\verb+lga_passengers+ & 0.561 & 0.620 & 0.704 & 0.563 & 0.606 & \textbf{0.892} & 0.526 & 0.537 & 0.366 & T & 0.592 & 0.537 & 0.674 & 0.535 & 0.774\\
\verb+measles+ & \textbf{0.947} & \textbf{0.947} & \textbf{0.947} & 0.486 & 0.118 & 0.103 & 0.281 & 0.153 & 0.391 & F/T & 0.030 & \textbf{0.947} & 0.041 & \textbf{0.947} & \textbf{0.947}\\
\verb+nile+ & \textbf{1.000} & \textbf{1.000} & \textbf{1.000} & 0.800 & \textbf{1.000} & \textbf{1.000} & 0.824 & \textbf{1.000} & 0.824 & 0.667 & \textbf{1.000} & \textbf{1.000} & \textbf{1.000} & 0.824 & \textbf{1.000}\\
\verb+ozone+ & 0.776 & 0.723 & 0.857 & 0.778 & 0.750 & \textbf{1.000} & 0.667 & \textbf{1.000} & 0.723 & 0.651 & 0.429 & \textbf{1.000} & 0.286 & 0.723 & \textbf{1.000}\\
\verb+quality_control_1+ & \textbf{1.000} & \textbf{1.000} & \textbf{1.000} & 0.667 & 0.667 & \textbf{1.000} & 0.667 & \textbf{1.000} & 0.500 & 0.286 & 0.667 & \textbf{1.000} & 0.667 & 0.667 & 0.800\\
\verb+quality_control_2+ & \textbf{1.000} & \textbf{1.000} & \textbf{1.000} & 0.667 & \textbf{1.000} & \textbf{1.000} & \textbf{1.000} & \textbf{1.000} & 0.750 & 0.429 & \textbf{1.000} & \textbf{1.000} & \textbf{1.000} & 0.750 & \textbf{1.000}\\
\verb+quality_control_3+ & \textbf{1.000} & \textbf{1.000} & \textbf{1.000} & 0.766 & 0.571 & \textbf{1.000} & \textbf{1.000} & \textbf{1.000} & 0.667 & T & 0.800 & \textbf{1.000} & \textbf{1.000} & 0.667 & \textbf{1.000}\\
\verb+quality_control_4+ & 0.810 & 0.873 & 0.787 & 0.561 & 0.658 & 0.787 & 0.658 & 0.780 & 0.780 & T & 0.241 & 0.780 & 0.608 & 0.780 & \textbf{0.909}\\
\verb+quality_control_5+ & \textbf{1.000} & \textbf{1.000} & \textbf{1.000} & 0.500 & \textbf{1.000} & \textbf{1.000} & \textbf{1.000} & \textbf{1.000} & \textbf{1.000} & 0.500 & \textbf{1.000} & \textbf{1.000} & \textbf{1.000} & \textbf{1.000} & \textbf{1.000}\\
\verb+rail_lines+ & 0.846 & 0.846 & \textbf{0.966} & 0.889 & \textbf{0.966} & \textbf{0.966} & 0.800 & 0.846 & 0.537 & 0.730 & 0.615 & 0.889 & 0.205 & 0.537 & \textbf{0.966}\\
\verb+ratner_stock+ & 0.776 & 0.824 & \textbf{0.868} & 0.559 & 0.396 & 0.776 & 0.754 & 0.824 & 0.280 & T & 0.203 & 0.824 & 0.378 & 0.571 & 0.824\\
\verb+robocalls+ & 0.800 & \textbf{0.966} & \textbf{0.966} & 0.750 & 0.862 & \textbf{0.966} & \textbf{0.966} & \textbf{0.966} & 0.636 & 0.846 & 0.714 & \textbf{0.966} & 0.714 & 0.636 & \textbf{0.966}\\
\verb+scanline_126007+ & 0.710 & 0.920 & \textbf{0.921} & 0.829 & 0.906 & 0.870 & 0.838 & 0.889 & 0.644 & T & 0.649 & 0.889 & 0.818 & 0.644 & 0.833\\
\verb+scanline_42049+ & 0.485 & 0.879 & \textbf{0.962} & 0.889 & 0.713 & 0.910 & 0.908 & 0.910 & 0.269 & T & 0.460 & 0.910 & 0.650 & 0.276 & 0.924\\
\verb+seatbelts+ & 0.824 & 0.838 & 0.683 & 0.583 & 0.735 & 0.683 & 0.621 & 0.683 & 0.452 & 0.383 & 0.563 & 0.735 & 0.583 & 0.621 & \textbf{0.974}\\
\verb+shanghai_license+ & \textbf{0.966} & 0.868 & 0.868 & 0.605 & 0.600 & 0.868 & 0.465 & 0.868 & 0.532 & 0.389 & 0.357 & 0.868 & 0.385 & 0.636 & 0.778\\
\verb+uk_coal_employ+ & M & M & M & 0.617 & M & 0.513 & 0.513 & M & 0.639 & M & M & M & M & 0.513 & \textbf{0.941}\\
\verb+unemployment_nl+ & 0.742 & \textbf{0.889} & 0.876 & 0.592 & 0.747 & 0.744 & 0.744 & 0.788 & 0.566 & F/T & 0.628 & 0.788 & 0.801 & 0.566 & 0.876\\
\verb+us_population+ & \textbf{1.000} & 0.889 & \textbf{1.000} & 0.615 & 0.232 & 0.471 & 0.276 & 0.500 & 0.159 & T & 0.889 & 0.889 & 0.113 & 0.889 & 0.889\\
\verb+usd_isk+ & \textbf{0.785} & 0.704 & \textbf{0.785} & 0.678 & 0.674 & \textbf{0.785} & 0.601 & 0.657 & 0.489 & 0.510 & 0.462 & 0.678 & 0.636 & 0.489 & \textbf{0.785}\\
\verb+well_log+ & 0.336 & 0.914 & 0.832 & 0.743 & 0.822 & \textbf{0.928} & 0.776 & 0.873 & 0.149 & T & 0.923 & 0.873 & 0.832 & 0.237 & 0.869\\
\hline
\end{tabular}
    }
    \label{tab:f1_uni}
\end{table*}

\begin{table*}[]
    \setlength{\tabcolsep}{4pt}
    \scriptsize
    \centering
        \caption{Cover for univariate datasets (\textit{best} mode). Values are unavailable either due to a failure (F), the method timing out (T) or missing
values in the series unsupported by the method (M). Highest values for each time series are shown in bold.}
    \scalebox{0.96}{
    \begin{tabular}{lcccccccccccccc|c}
Dataset & \textsc{amoc} & \textsc{binseg} & \textsc{bocpd} & \textsc{bocpdms} & \textsc{cpnp} & \textsc{ecp} & \textsc{kcpa} & \textsc{pelt} & \textsc{prophet} & \textsc{rbocpdms} & \textsc{rfpop} & \textsc{segneigh} & \textsc{wbs} & \textsc{zero} & \textsc{watch}\\
\hline
\verb+bank+ & \textbf{1.000} & \textbf{1.000} & \textbf{1.000} & 0.997 & 0.103 & 0.238 & 0.509 & 0.509 & \textbf{1.000} & T & 0.036 & \textbf{1.000} & 0.053 & \textbf{1.000} & \textbf{1.000}\\
\verb+bitcoin+ & 0.771 & 0.771 & \textbf{0.822} & 0.773 & 0.364 & 0.772 & 0.778 & 0.794 & 0.723 & T & 0.168 & 0.771 & 0.409 & 0.516 & 0.816\\
\verb+brent_spot+ & 0.503 & 0.650 & \textbf{0.667} & 0.265 & 0.437 & 0.653 & 0.571 & 0.659 & 0.527 & T & 0.235 & 0.659 & 0.310 & 0.266 & 0.652\\
\verb+businv+ & 0.574 & 0.562 & \textbf{0.693} & 0.459 & 0.402 & 0.690 & 0.405 & 0.647 & 0.539 & 0.459 & 0.123 & 0.647 & 0.154 & 0.461 & 0.554\\
\verb+centralia+ & 0.675 & 0.675 & \textbf{0.753} & 0.650 & 0.675 & \textbf{0.753} & 0.675 & 0.675 & 0.675 & 0.624 & 0.568 & 0.675 & 0.253 & 0.675 & \textbf{0.753}\\
\verb+children_per_woman+ & 0.804 & 0.798 & 0.801 & 0.427 & 0.486 & 0.718 & 0.613 & 0.798 & 0.521 & 0.715 & 0.154 & 0.798 & 0.284 & 0.429 & \textbf{0.868}\\
\verb+co2_canada+ & 0.527 & 0.747 & \textbf{0.773} & 0.584 & 0.528 & 0.751 & 0.739 & 0.705 & 0.605 & 0.617 & 0.497 & 0.752 & 0.552 & 0.278 & 0.742\\
\verb+construction+ & 0.629 & 0.575 & 0.585 & 0.571 & 0.375 & 0.524 & 0.395 & 0.423 & 0.561 & 0.575 & 0.154 & 0.575 & 0.300 & 0.575 & \textbf{0.735}\\
\verb+debt_ireland+ & 0.635 & 0.717 & 0.688 & 0.729 & 0.777 & 0.798 & 0.747 & 0.777 & 0.321 & 0.396 & 0.489 & 0.777 & 0.248 & 0.321 & \textbf{0.832}\\
\verb+gdp_argentina+ & \textbf{0.737} & \textbf{0.737} & \textbf{0.737} & 0.711 & 0.461 & \textbf{0.737} & 0.367 & 0.667 & 0.592 & 0.711 & 0.346 & \textbf{0.737} & 0.326 & \textbf{0.737} & \textbf{0.737}\\
\verb+gdp_croatia+ & 0.708 & 0.708 & 0.708 & 0.675 & 0.708 & 0.708 & 0.623 & 0.708 & 0.708 & 0.708 & 0.353 & 0.708 & 0.108 & 0.708 & \textbf{0.773}\\
\verb+gdp_iran+ & 0.583 & 0.583 & 0.583 & 0.611 & 0.448 & 0.583 & 0.505 & 0.505 & 0.583 & 0.692 & 0.248 & 0.583 & 0.295 & 0.583 & \textbf{0.730}\\
\verb+gdp_japan+ & \textbf{0.802} & \textbf{0.802} & \textbf{0.802} & 0.777 & 0.522 & \textbf{0.802} & 0.525 & \textbf{0.802} & \textbf{0.802} & \textbf{0.802} & 0.283 & \textbf{0.802} & 0.283 & \textbf{0.802} & \textbf{0.802}\\
\verb+global_co2+ & \textbf{0.758} & \textbf{0.758} & \textbf{0.758} & 0.745 & 0.381 & 0.665 & 0.602 & 0.743 & 0.284 & 0.745 & 0.368 & \textbf{0.758} & 0.338 & \textbf{0.758} & \textbf{0.758}\\
\verb+homeruns+ & \textbf{0.694} & 0.683 & \textbf{0.694} & 0.681 & 0.537 & \textbf{0.694} & 0.501 & 0.683 & 0.575 & 0.506 & 0.392 & 0.683 & 0.407 & 0.511 & \textbf{0.694}\\
\verb+iceland_tourism+ & \textbf{0.946} & \textbf{0.946} & \textbf{0.946} & 0.936 & 0.393 & 0.842 & 0.655 & 0.830 & 0.498 & 0.936 & 0.293 & \textbf{0.946} & 0.512 & \textbf{0.946} & \textbf{0.946}\\
\verb+jfk_passengers+ & 0.844 & 0.839 & 0.837 & \textbf{0.855} & 0.583 & 0.807 & 0.563 & 0.839 & 0.373 & T & 0.393 & 0.839 & 0.514 & 0.630 & 0.812\\
\verb+lga_passengers+ & 0.434 & 0.541 & 0.547 & 0.475 & 0.478 & \textbf{0.653} & 0.536 & 0.534 & 0.446 & T & 0.426 & 0.543 & 0.501 & 0.383 & 0.604\\
\verb+measles+ & \textbf{0.951} & \textbf{0.951} & \textbf{0.951} & 0.950 & 0.098 & 0.105 & 0.400 & 0.232 & 0.616 & F/T & 0.046 & \textbf{0.951} & 0.084 & \textbf{0.951} & \textbf{0.951}\\
\verb+nile+ & \textbf{0.888} & 0.880 & \textbf{0.888} & 0.870 & 0.880 & \textbf{0.888} & 0.758 & 0.880 & 0.758 & 0.876 & 0.880 & 0.880 & 0.880 & 0.758 & \textbf{0.888}\\
\verb+ozone+ & 0.721 & 0.600 & 0.602 & 0.735 & 0.458 & 0.676 & 0.451 & 0.635 & 0.574 & 0.755 & 0.358 & 0.627 & 0.309 & 0.574 & \textbf{0.822}\\
\verb+quality_control_1+ & \textbf{0.996} & 0.992 & \textbf{0.996} & 0.990 & 0.721 & 0.989 & 0.620 & 0.992 & 0.693 & 0.780 & 0.706 & 0.992 & 0.687 & 0.503 & 0.882\\
\verb+quality_control_2+ & 0.927 & 0.922 & \textbf{0.927} & 0.921 & 0.922 & 0.927 & \textbf{0.927} & 0.922 & 0.723 & 0.637 & 0.922 & 0.922 & 0.922 & 0.638 & 0.923\\
\verb+quality_control_3+ & \textbf{0.997} & 0.996 & \textbf{0.997} & 0.987 & 0.831 & \textbf{0.997} & \textbf{0.997} & 0.996 & 0.500 & T & 0.978 & 0.996 & 0.996 & 0.500 & 0.993\\
\verb+quality_control_4+ & \textbf{0.742} & 0.673 & 0.673 & 0.670 & 0.508 & 0.540 & 0.535 & 0.673 & 0.673 & T & 0.077 & 0.673 & 0.506 & 0.673 & 0.673\\
\verb+quality_control_5+ & \textbf{1.000} & \textbf{1.000} & \textbf{1.000} & 0.994 & \textbf{1.000} & \textbf{1.000} & \textbf{1.000} & \textbf{1.000} & \textbf{1.000} & 0.994 & \textbf{1.000} & \textbf{1.000} & \textbf{1.000} & \textbf{1.000} & \textbf{1.000}\\
\verb+rail_lines+ & 0.786 & 0.786 & 0.768 & 0.769 & \textbf{0.786} & 0.768 & 0.773 & 0.786 & 0.534 & 0.769 & 0.482 & 0.786 & 0.103 & 0.428 & 0.784\\
\verb+ratner_stock+ & 0.874 & 0.914 & 0.906 & 0.872 & 0.382 & 0.874 & 0.771 & \textbf{0.914} & 0.444 & T & 0.162 & \textbf{0.914} & 0.182 & 0.450 & 0.906\\
\verb+robocalls+ & 0.666 & 0.788 & \textbf{0.808} & 0.741 & 0.677 & \textbf{0.808} & \textbf{0.808} & 0.760 & 0.601 & 0.682 & 0.569 & 0.760 & 0.559 & 0.601 & \textbf{0.808}\\
\verb+scanline_126007+ & 0.634 & 0.633 & 0.631 & \textbf{0.677} & 0.433 & 0.390 & 0.494 & 0.444 & 0.503 & T & 0.316 & 0.633 & 0.329 & 0.503 & 0.523\\
\verb+scanline_42049+ & 0.425 & 0.861 & \textbf{0.892} & 0.875 & 0.577 & 0.866 & 0.860 & 0.862 & 0.441 & T & 0.257 & 0.862 & 0.690 & 0.211 & 0.874\\
\verb+seatbelts+ & 0.683 & 0.797 & 0.800 & 0.630 & 0.809 & 0.800 & 0.702 & 0.797 & 0.635 & 0.533 & 0.660 & 0.797 & 0.727 & 0.528 & \textbf{0.832}\\
\verb+shanghai_license+ & \textbf{0.930} & 0.920 & 0.920 & 0.911 & 0.474 & 0.920 & 0.497 & 0.920 & 0.804 & 0.763 & 0.381 & 0.920 & 0.351 & 0.547 & 0.847\\
\verb+uk_coal_employ+ & M & M & M & 0.504 & M & 0.356 & 0.356 & M & 0.481 & M & M & M & M & 0.356 & \textbf{0.650}\\
\verb+unemployment_nl+ & 0.627 & 0.669 & \textbf{0.669} & 0.572 & 0.503 & 0.515 & 0.476 & 0.650 & 0.507 & F/T & 0.246 & 0.650 & 0.447 & 0.507 & 0.651\\
\verb+us_population+ & \textbf{0.803} & \textbf{0.803} & 0.737 & 0.801 & 0.391 & 0.508 & 0.271 & 0.506 & 0.135 & T & \textbf{0.803} & \textbf{0.803} & 0.050 & \textbf{0.803} & \textbf{0.803}\\
\verb+usd_isk+ & 0.865 & 0.764 & 0.853 & \textbf{0.866} & 0.540 & 0.858 & 0.714 & 0.791 & 0.436 & 0.770 & 0.166 & 0.791 & 0.401 & 0.436 & 0.858\\
\verb+well_log+ & 0.463 & 0.828 & 0.793 & 0.769 & 0.779 & \textbf{0.835} & 0.798 & 0.782 & 0.411 & T & 0.787 & 0.782 & 0.768 & 0.225 & 0.798\\
\hline
\end{tabular}

    }
    \label{tab:cover_uni}
\end{table*}

\begin{figure*}[]
    \centering
    \includegraphics[width=0.9\textwidth]{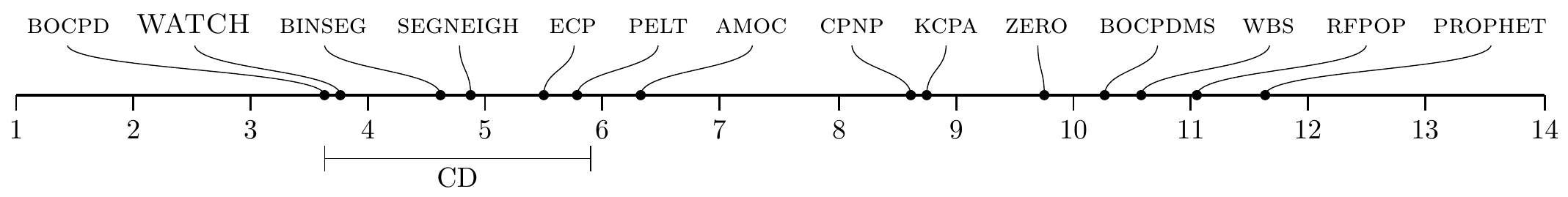}
    \caption{F1 rank plot for univariate data in the \textit{best} mode.}
\label{fig:rankplot_f1_uni}

    \centering
    \includegraphics[width=0.9\textwidth]{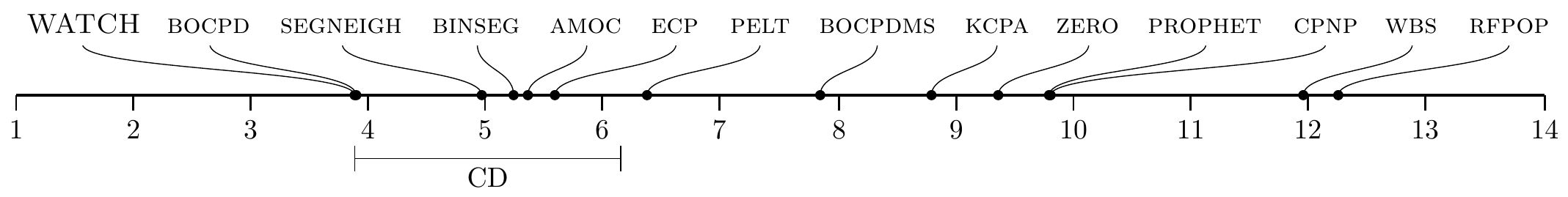}
    \caption{Cover rank plot for univariate data in the \textit{best} mode.}
\label{fig:rankplot_cover_uni}
\end{figure*}

The results of \WATCH{} for univariate datasets strongly suggest that the method may be successfully applied not only for multivariate (and particularly high-dimensional) but also for univariate time series datasets. The rank plots for univariate datasets reported in Figure \ref{fig:rankplot_f1_uni} and \ref{fig:rankplot_cover_uni} show that, overall, \WATCH{} positions itself as the best (in terms of Cover) and second best (in terms of F1) method and significantly outperforms most of the other competitors. This result validates our considerations made inspecting the numerical values of the metrics in Tables \ref{tab:f1_uni} and \ref{tab:cover_uni}.

\section{Conclusion}
\label{sec:conclusion}

In this work we addressed the change point detection problem focusing on multivariate and high-dimensional data, which represents a challenging scenario for most of state-of-the-art methods. We proposed \WATCH{}, a novel Wasserstein distance-based change point detection approach that models the current data distribution and monitors its behavior over time. A threshold-ratio based approach is used to adaptively detect change points in dynamically changing time series data. An extensive experimental evaluation with a large number of benchmark datasets reveals the potential of our method, which outperforms all considered competitors in the majority of the cases. A statistical evaluation was performed to validate the obtained results. As future work, we aim to investigate possible extensions of our method to leverage periodic behavior in time series data, as well as possible applications in lifelong learning settings.


\section*{Acknowledgments}
The work presented in this article was partially funded by DARPA through the project ``Lifelong Streaming Anomaly Detection'' (Grant N. A19-0131-003 and A21-0113-002). We also acknowledge the support of NVIDIA through the donation of a Titan V GPU.

\bibliographystyle{IEEEtran}
\bibliography{references}

\end{document}